# UM_FHS at TREC 2024 PLABA: Exploration of Fine-tuning and AI agent approach for plain language adaptations of biomedical text

**Primoz Kocbek*,1,2, Leon Kopitar1,3, Zhihong Zhang4, Emirhan Aydın5, Maxim Topaz4, Gregor Stiglic1,3,6**
[1] University of Maribor, Faculty of Health Sciences [2] University of Ljubljana, Faculty of Medicine [3] University of Maribor, Faculty of Electrical Engineering and Computer Science [4] Columbia University, School of Nursing [5] Manisa Celal Bayar University [6] University of Edinburgh, Usher Institute
*Corresponding Author*

## Abstract

This paper describes our submissions to the TREC 2024 PLABA track with the aim to simplify biomedical abstracts for a K8-level audience (13–14 years old students). We tested three approaches using OpenAI's gpt-4o and gpt-4o-mini models: baseline prompt engineering, a two-AI agent approach, and fine-tuning. Adaptations were evaluated using qualitative metrics (5-point Likert scales for simplicity, accuracy, completeness, and brevity) and quantitative readability scores (Flesch-Kincaid grade level, SMOG Index). Results indicated that the two-agent approach and baseline prompt engineering with gpt-4o-mini models show superior qualitative performance, while fine-tuned models excelled in accuracy and completeness but were less simple. The evaluation results demonstrated that prompt engineering with gpt-4o-mini outperforms iterative improvement strategies via two-agent approach as well as fine-tuning with gpt-4o. We intend to expand our investigation of the results and explore advanced evaluations.

## 1  Introduction

This paper presents an overview of the *um_fhs'* submissions to the TREC 2024 Plain Language Adaptation of Biomedical Abstracts (PLABA) track[1], more specifically Task 2 for complete abstract adaptations that consists of end-to-end biomedical abstracts adaptations for the general public using plain language. The guidelines assume that the average literacy level is lower than grade 8 (<K8), i.e. students 13–14 years old, as is recommended by National Institutes of Health (NIH) for written health materials (Hutchinson et al., 2016).

The *um_fhs* team consisted of researchers from University of Maribor, Faculty of Health Sciences, Slovenia and collaborators from Columbia University, School of Nursing, U.S. as well as Manisa Celal Bayar University, Turkiye. Our research is a continuation of previous research in the field of healthcare, such as our focus on different summarization tasks (Kocbek et al., 2022; Stiglic et al., 2022) and general application of Large Language Models (LLMs) on different downstream tasks in healthcare (Kopitar et al., 2024; Kocbek et al., 2023).

Our submissions used the most used LLMs from OpenAI at the time of submission, specifically `gpt-4o` and `gpt-4o-mini` models, where we created three approaches per model through their API. Note that we used the OpenAI API with a signed Data Processing Addendum (DPA), which ensures GDPR compliance[2].

Firstly, we used a no-context approach with only prompt engineering. Next, we expanded the approach to include two AI agents improving output iteratively and lastly, we fine-tuned (FT) the training data on both models.

Both qualitative and quantitative metrics were used to evaluate end-to-end adaptations. The gold standard for such evaluations is often post-hoc human (qualitative) evaluation of domain experts, such as using 5-, 6- or 10- point Likert scales for concepts like sentence simplicity, term simplicity, term accuracy, fluency, completeness, faithfulness[1]. We used a 5-point Likert scale for simplicity, i.e. outputs should be easy to understand, accuracy, i.e. outputs should contain the accurate information, completeness, i.e. outputs should seek to minimize information lost from the original text and brevity, i.e. outputs should be concise[1]. For quantitative metrics, we used a basic readability scores: Flesch–Kincaid (FK) grade level (Flesch, 1948) and SMOG (Simple Measure of Gobbledygook) (Mc Laughlin, 1969). The second is often recommended for healthcare, where in both cases a score below 8 indicates <K8. We evaluated the output according to results of both types of metrics and selected three submissions for the track.

---

[1] https://bionlp.nlm.nih.gov/plaba2024/

[2] https://openai.com/security-and-privacy/

In the following section we detail the methods, provide results of experiment and their evaluation as well as the selection for the three submissions from our group to TREC 2024 PLABA, Task 2.

## 2 Methodology

### 2.1 Data preparation

The public PLABA dataset (Attal et al., 2023) was used as a reference training dataset, consisting of 750 manually adapted abstracts that are both document- and sentence-aligned. The format of the dataset was a `.json` file, where Python script was used to extract the list of abstract sentences together with their adaptations for input/processing in our approaches together with the questions, since the abstracts have been retrieved to answer consumer questions asked on MedlinePlus (Attal et al., 2023). Our approaches took a list of sentences as input and produced a list of adapted sentences as output.

The PLABA training dataset consisted of 750 abstracts with adaptations, where 10 abstracts were retrieved to answer a question. For each such question we split the data 80% for training (733 samples) and 20% for validation (184 samples) with respect to 'pmid' since the abstract could have more than one adaptation. We used this split for all initial model evaluations.

### 2.2 Prompt Engineering

A system prompt was created (Appendix A), where the task, tone, input and output were defined in general. Next, we created the baseline prompt (Appendix B), where we stated in more detail the task together with the modified guidelines for annotators for PLABA [3], where we manually modified the text for LLM ingestion (Appendix C). Our next step was to create prompts for the interaction of two AI agents (Appendix D), more specifically we created a discussion, called a `thread`, for each input, where the first agent created an adaptation using the baseline prompt and the second AI agent in the persona of a "smart 13-14 year old student" asked clarification questions and then first AI agent modified the adaptation using relevant answers to the questions from the second AI agent. Additionally, we used FT where we used the system prompt (Appendix A).

The above process for the described approaches relied heavily on engineering optimal prompts, which were iteratively improved. The initial prompts consisted of general prompt templates, which were that modified using a trial-and-error approach. For example, we needed to explicitly state that the length of the list of sentences in the input has to be the same in the output, that a sentence can be omitted if the information is not relevant to the guidelines, and emphasize that medical terms needs to be explained in a plain language.

### 2.3 Models

Since the training dataset is public, we additionally used a GDPR compliant version of OpenAI API, more specifically we used gpt-4o-mini, model version `gpt-4o-mini-2024-07-18` and gpt-4o, model version `gpt-4o-2024-08-06`.

As described in section 2.2, we created three approaches for each model: baseline (1), two AI agents (2) and FT models (3). Note that FT gpt-4o and gpt-4o-mini models became available on Aug 20th 2024[4]. FT on gpt-4o and gpt-4o-mini was performed on 25th August 2024, with the following hyperparameters: epochs 3, batch size 1, LR multiplier 2, random seed 741667963. The training and validation split was 80:20 (see section 2.1). The results from FT were follows: `gpt-4o` training loss 1.099, full validation loss 0.8336 and `gpt-4o-mini` training loss 1.0489, full validation loss 0.967. Note that the difference between training and full validation loss is less for gpt-4o-mini FT than for gpt-4o FT.

We produced six end-to-end adaptations for PLABA Task 2, we refer to them as `gpt-4o_baseline`, `gpt-4o_two_agents`, `gpt-4o-`

---

[3] https://bionlp.nlm.nih.gov/plaba2024/annotation_guidelines.pdf

[4] https://openai.com/index/gpt-4o-fine-tuning/

ft, gpt-4o-mini_baseline, gpt-4o-mini_two_agents, gpt-4o-mini-ft in the rest of the paper.

## 2.4 Evaluation Metrics

Two basic quantitative metrics or readability scores were used for evaluating the outputs, FK grade level and SMOG, both using US grade level based, i.e. average literacy level should be lower than grade 8 (<K8) or students 13–14 years old and therefore the average results should be below (Hutchinson et al., 2016). However, this target can be misleading, since it turned out that the ground truth in the training dataset was higher than expected with a mean of 11.64 (SD=2.43) and 13.8 (SD=2.18) for SMOG (Table 1), both indicating K12 or even K14 grade level. Additionally, since for SMOG texts of fewer than 30 sentences are statistically invalid as the formula was normed on 30-sentence samples[5] and our abstract were the length around 11, we therefore primarily use FK grade level results. We also used appropriate statistical tests to compare experimental results with ground truth results.

For qualitative evaluation we used a randomly selected sample (n=40) from text set (N=400) for manual evaluations, using a 5-point Likert scale for simplicity, accuracy, completeness and brevity were used.

## 3 Experiments and Evaluations

We used 20% (n=184) of the training data in FT validation, hence this data was not directly used by any of our approaches, therefore we used it to compare the results of our approaches. Firstly, we looked at boxplots for visually comparison FK grade level and SMOG to ground truths, i.e. the adaptations of the training set (Figure 1). We can observe ground truth FK grade levels and SMOG index means were well above the hypothetical grade level 8 with 11.64 (SD=2.43) and 13.8 (SD=2.18) respectively. The baseline and two agent approaches for gpt-4o were the only approaches below the mean of 8, however it turns out that this was a negative when qualitatively evaluating the test set for completeness (Table 2). The FT approaches were the closest to the ground truth.

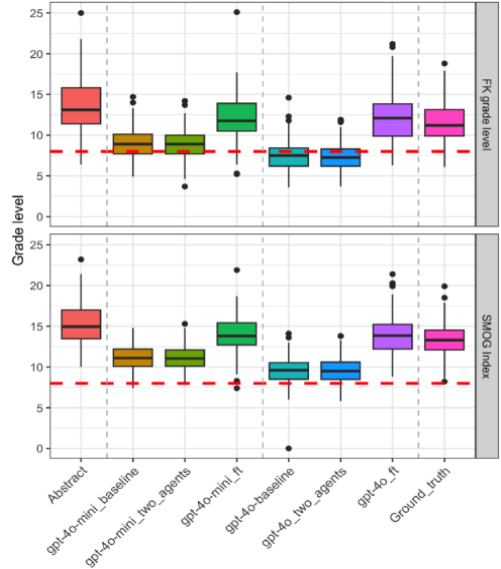

Figure 1: Visual comparison on grade levels (FK grade level and SMOG Index) with Abstract, gpt-4o and gpt-4o-mini approaches and ground truth on the training validation set. The red dashed line marks the hypothetical 8th grade line.

We performed a paired t-test comparing ground truth with all other approaches, where we observed a statistical difference in all cases (Table 1).

|  | FK grade level | | SMOG Index | |
|---|---|---|---|---|
|  | M (SD) | p-value | M (SD) | p-value |
| **Abstract** | 13.67 (SD=3.29) | <0.001 | 15.3 (SD=2.63) | <0.001 |
| **gpt-4o-mini_ baseline** | 8.93 (SD=1.76) | <0.001 | 11.12 (SD=1.49) | <0.001 |
| **gpt-4o-mini_ two_agents** | 8.91 (SD=1.85) | <0.001 | 11.11 (SD=1.52) | <0.001 |
| **gpt-4o-mini_ ft** | 12.16 (SD=2.7) | 0.0054 | 13.98 (SD=2.1) | <0.001 |
| **gpt-4o- baseline** | 7.4 (SD=1.72) | <0.001 | 9.54 (SD=1.67) | <0.001 |
| **gpt-4o_ two_agents** | 7.29 (SD=1.59) | <0.001 | 9.6 (SD=1.63) | <0.001 |
| **gpt-4o_ ft** | 12.2 (SD=2.76) | 0.0054 | 13.89 (SD=2.31) | <0.001 |
| **Ground Truth** | 11.64 (SD=2.43) | / | 13.35 (SD=1.95) | / |

Table 1: Statistical comparison of grade levels (FK grade level and SMOG Index) from Ground Truth to Abstract, gpt-4o and gpt-4o-mini approaches on the training validation set.

---

[5] https://en.wikipedia.org/wiki/SMOG

|  | Simplicity | Accuracy | Completeness | Brevity | Total Score |
|---|---|---|---|---|---|
| gpt-4o-mini_baseline | 4.08 (SD=1.02) | 4.2 (SD=0.88) | 4.42 (SD=0.75) | 4.03 (SD=0.77) | 16.73 (SD=2.49) |
| gpt-4o-mini_two_agents | 4.22 (SD=0.95) | 4.25 (SD=0.87) | 4.38 (SD=0.7) | 4.08 (SD=0.89) | 16.92 (SD=2.79) |
| gpt-4o-mini_ft | 3.75 (SD=0.93) | 4.1 (SD=0.9) | 4.3 (SD=0.79) | 3.6 (SD=0.9) | 15.75 (SD=2.27) |
| gpt-4o-baseline | 4.32 (SD=0.73) | 3.88 (SD=0.61) | 3.45 (SD=0.71) | 4.28 (SD=0.82) | 15.93 (SD=1.8) |
| gpt-4o_two_agents | 4.45 (SD=0.71) | 3.62 (SD=0.9) | 3.7 (SD=0.88) | 4.25 (SD=0.81) | 16.02 (SD=2.29) |
| gpt-4o_ft | 3.8 (SD=0.76) | 4.2 (SD=0.76) | 4.3 (SD=0.72) | 3.48 (SD=0.72) | 15.78 (SD=2.02) |

Table 2: Qualitative evaluation of a sample (n=40) test set for simplicity, accuracy, completeness and brevity.

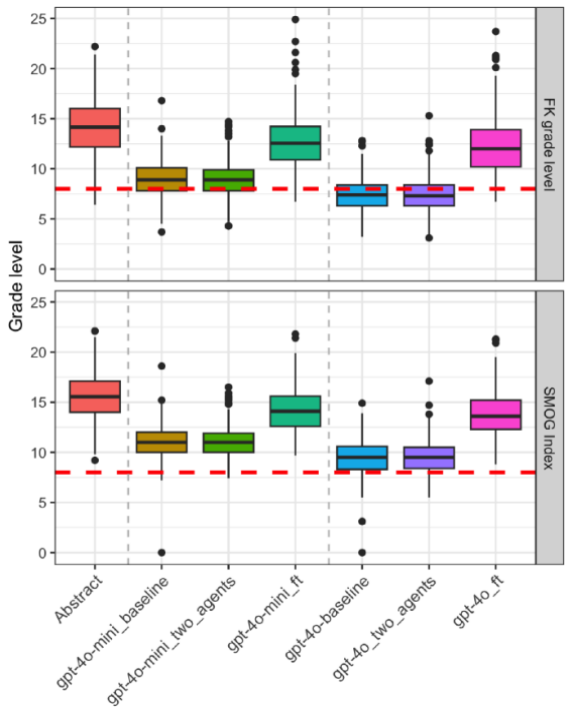

Figure 2: Visual comparison on grade levels (FK grade level and SMOG Index) with Abstract, gpt-4o and gpt-4o-mini approaches on the test set. The red dashed line marks the hypothetical 8th grade line.

Next, we visually compared (Figure 1, Figure 2) the training validation data (n=184) and test data (n=400), where the graphs look quite similar apart from a higher number of high grade level outliers in the `gpt-4o-mini` AI agents, FT approaches, as well as the `gpt-4o` FT approach.

We also performed a qualitative evaluation (5-point Likert scale) by healthcare experts on a sample (n=40) of the test set (N=400) for simplicity, accuracy, completeness and brevity. We observed (Table 2) that both FT models scored last for simplicity, `gpt-4o-mini_ft` with 3.75 (SD=0.93) and `gpt-4o_ft` with 3.8 (SD=0.76). For accuracy and completeness it is interesting that all `gpt-4o-mini` models performed well as did `gpt-4o_ft`. For brevity `gpt-4o_baseline` and `gpt-4o_two_agent` aproaches perfomed best with 4.28 (SD=0.82) and 4.25 (SD=0.81).

When considering the aformention scores we concluded that `gpt-4o-mini_baseline` and `gpt-4o-mini_two agent` approaches perfomed best, wheras the other seem to be quite close together (Table 2).

Considering mostly the qualitative results (Table 2, Figure 2), we selected `gpt-4o-mini_two_agent` and `gpt-4o-mini_baseline` as the top two submissions. For the third submission, we selected `gpt-4o_ft`, since it is our view that accuracy and completeness outweigh the other two evaluation metrics. For example, a higher mean FK grade level in ground truths averaging 11.64 (SD=2.43), might indicate that completeness and accuracy are quite important.

## 4 Official Ranking at PLABA 2024

The official ranking for our runs of Task 2 (end-to-end adaptations) are shown in Table 3. The runs were manually evaluated by NIST evaluators on 3-point Likert scale (-1, 0, 1) for each sentence, The average of the 4 categories was mapped to [0,1]. Our first submission (sub1) `gpt-4o-mini_two agent` compared to our second submission (sub2) `gpt-4o-mini_baseline`, interestingly performed worse in all four categories, especially for accuracy and simplicity underperforming by almost 5%. Our third submission (sub3) `gpt-4o_ft` performed just slightly (1.5%) worse than our second submission (sub 2), underperforming in simplicity, almost 5%, and brevity, almost 9%.

It is somewhat surprising that a baseline prompt outperformed `gpt-4o-mini_baseline` our iterative improvement strategy in `gpt-4o-mini_two agent` and intend to investigate it further.

| Run | Accuracy | Completeness | Simplicity | Brevity | Final Avg. |
|---|---|---|---|---|---|
| GPT | 0.9307 | 0.8118 | 0.9232 | 0.8755 | 0.8853 |
| plaba_um_fhs_sub2 | 0.9447 | 0.8588 | 0.8909 | 0.7930 | 0.8718 |
| task2_moa_tier3_post | 0.9395 | 0.8681 | 0.8860 | 0.6586 | 0.8381 |
| plaba_um_fhs_sub1 | 0.8985 | 0.8427 | 0.8421 | 0.7618 | 0.8363 |
| task2_moa_tier2_post | 0.9307 | 0.8537 | 0.8642 | 0.6468 | 0.8238 |
| TREC2024_SIB_run1 | 0.9374 | 0.8614 | 0.8363 | 0.6594 | 0.8236 |
| plaba_um_fhs_sub3 | 0.9504 | 0.8681 | 0.7949 | 0.6728 | 0.8215 |
| gpt35_dspy | 0.9117 | 0.8700 | 0.7654 | 0.6533 | 0.8001 |
| task2_moa_tier1_post | 0.8948 | 0.8463 | 0.8197 | 0.6272 | 0.7970 |
| TREC2024_SIB_run3 | 0.8170 | 0.7324 | 0.8756 | 0.7392 | 0.7910 |
| TREC2024_SIB_run4 | 0.9433 | 0.8891 | 0.6965 | 0.6210 | 0.7875 |
| UAms-ConBART-Cochrane | 0.9546 | 0.9238 | 0.5653 | 0.6781 | 0.7804 |
| UAms-BART-Cochrane | 0.9772 | 0.9466 | 0.4881 | 0.6906 | 0.7756 |
| LLaMA-8B-4bit-MedicalAbstract-seq-to-seq-v1 | 0.8831 | 0.8447 | 0.7792 | 0.5240 | 0.7578 |
| gpt-final | 0.9537 | 0.8996 | 0.7850 | 0.3782 | 0.7541 |
| mistral-fix | 0.8818 | 0.8603 | 0.7118 | 0.4155 | 0.7174 |
| mistral-FINAL | 0.8788 | 0.8567 | 0.7188 | 0.4122 | 0.7166 |
| LLaMa_3.1_70B_instruction_2nd_run | 0.7440 | 0.7725 | 0.7044 | 0.5903 | 0.7028 |
| bart_base_ft | 0.8424 | 0.7132 | 0.4500 | 0.4177 | 0.6058 |

Table 3: PLABA 2024 Task 2 full results evaluated for simplicity, accuracy, completeness, brevity and final average.

## 5 Conclusions and Future Work

Our approach for the TREC 2024 PLABA track end-to-end adaptations biomedical abstracts for the general public using plain language (Task 2) used multiple approaches from the currently most commonly used LLMs, `gpt-4o` and `gpt-4o-mini`. We used simple prompt engineering, two AI agents who iteratively improved output and a FT approach.

Since it is expected that the new upcoming LLMs will be more capable than the current generation, our focus was more on the evaluation side and selection of the most appropriate approach. We looked at some classic readability scores, such as FK grade level and SMOG index, focusing on the former since the latter should be more appropriate on longer texts. Qualitative metrics remain the gold standard and for the selection of the submissions we largely relied on those metrics.

For future research direction, we intend to focus on more complex quantitative metrics (semantic similarity and LLM based), use the results here for a baseline comparison with open source local LLMs, such as Llama 3.3, and look for signs of benchmark contamination in newer LLMs.

**Limitations**

Some of the limitations:

- Only two proprietary OpenAI models were used, for private healthcare data a locally deployed LLMs should be considered/used.

- Only two quantitative metrics were used in evaluation, semantic similarity metrics such as BERTScore and LLM-based metrics (GPTScore, G-Eval) were not used.

- Small sample dataset from curated biomedical abstracts might differ from real world long complex medical texts written by physicians or other healthcare experts.

- Benchmark contamination analysis was not performed, there are signs that newer LLMs are contaminated with public benchmark datasets (Ahuja et al., 2024).

- Readability metrics may not fully capture the nuances of language and the specific needs of the target audience.

- Manual evaluations using Likert point scales may be subject to bias and inconsistency in ratings.

## Acknowledgments

This work was funded by the European Union under the Horizon Europe grant #101159018.

## A  System prompt

```
You are tasked with adapting a list of sentences from a medical text into
plain language, suitable for readers at a K8 level (13 to 14 years old). The
adaptations must be simple, accurate, complete, brief, and fluent, ensuring
that the reader fully understands the content. Rules for splitting complex
sentences, substituting medical jargon with common alternatives, and
explaining terms with no substitutions. The tone should be casual and
understandable. You will be given a list of sentences, formatted as
["SENTENCE_1", "SENTENCE_2",...,"SENTENCE_N"]. Return only the list of
adaptations as ["ADAPTATION_1", "ADAPTATION_2",...,"ADAPTATION_N"]. Double-
check your work to ensure that the adaptations follow the guidelines, are as
complete as possible, are the same length as the list of sentences, and omit
any unnecessary information.
```

## B  Baseline prompt

```
You are tasked with adapting a list of sentences from a medical text into
plain language, suitable for readers at a K8 level (13 to 14 years old). The
adaptations must be simple, accurate, complete, brief, and fluent, ensuring
that the reader fully understands the content. Use the adaptation guidelines
provided below ##Adaptation guidelines##, which include rules for splitting
complex sentences, substituting medical jargon with common alternatives, and
explaining terms with no substitutions. The tone should be casual and
understandable.
You will be given a list of sentences, formatted as ["SENTENCE_1",
"SENTENCE_2",...,"SENTENCE_N"]. Your goal is to return a list of adapted
sentences in the same format, where each adapted sentence corresponds to the
original sentence at the same position in the list.
Prioritize making the adaptation as complete as possible to ensure full
understanding by a K8-level reader.
Split sentences when necessary to improve clarity.
If a sentence is already simple and understandable, you may carry it over
without changes.
If a sentence is irrelevant to consumer understanding, you should omit it and
replace it with "" in the output list.
If a sentence can be made simpler or clearer by replacing technical terms, do
so according to the guidelines.
Ensure that no information from other source sentences is merged into the
adaptation of any given sentence.
Output only the list of adaptations as ["ADAPTATION_1",
"ADAPTATION_2",...,"ADAPTATION_N"]. Double-check your work to ensure that the
adaptations follow the guidelines, are as complete as possible, are the same
length as the list of sentences, and omit any unnecessary information.
```

## C  Modified guidelines

```
These are guidelines for plain text adaptation from medical texts. The
guidelines also feature level of importance for specific concepts, if a word
or multiple words are encased "", that means that this concept has the highest
priority concept and should always be adhered to in plain language adaptations,
if a word or multiple words are encased in || that means a very high priority
concept and should be adhered to in plain language adaptations except if it
contradicts with a "" concept. Similarly word or multiple words encased between
[] are high priority concepts and should be adhered to except if it contradicts
"" or []. Examples sentences or example words for plain language adaptations
are provide in the format // // -> // //, where the first in  // // is the
original and second sentence  in // // the plain language adaptation.
```

Education level of audience for adapted (target) text: "K8 (8th grade level students, schooling age 13 to 14)"

|Splitting sentences|: if a sentence is long and contains two or more complete thoughts, it should be split into multiple sentences that are simpler. All such sentences will be entered in the same cell to the right of the source sentence, separating them with periods as per usual.

|Carrying over sentences or phrases|: a sentence or phrase need not be paraphrased if it is already understandable for consumers; it can simply be carried over as is. Similarly, some sentences may only need one or two terms to be substituted, but no syntactic changes made.

|Ignoring sentences|: if a source sentence is not relevant to consumer understanding of the document, it should be ignored, and the cell to the right of it left blank, for example:
1) Sentences that expound on experimental procedures not relevant to conclusions, such as 'Blood pressure of study participants was measured in mmHg using a sphygmomanometer.',
2) Adapt (do not ignore) sentences mentioning or implying that "Future studies are needed for this topic..."

|Resolving anaphora|: if pronouns in the source sentence refer to something in the previous sentence that is necessary for understanding the current, replace them with their referents in the target sentence. For example: //Cardiovascular disease is the leading cause of mortality.// -> //Heart disease is the leading cause of death.//, //It is influenced by genetics as well as lifestyle.// -> //Heart disease is influenced by heredity and lifestyle.//

General guidelines:
1) [Change passive voice to active voice when possible.] Example //A total of 24 papers were reviewed// -> //We reviewed a total of 24 papers//,
2) [If a source sentence contains a subheading, such as Background:, Results:,]
a) [And is followed by a complete sentence, omit the subheadings, such as Background:, Results: in the target text], example //Objective: Our aim is to evaluate management of foreign bodies in the upper gastrointestinal tract.// -> //Our aim is to rate treatment of foreign objects stuck in the upper digestive tract.// b) [And is followed by an incomplete sentence, convert the partial or incomplete sentence to a complete target sentence by folding in the subheading based on context], examples //Objective: To evaluate management of foreign bodies in the upper gastrointestinal tract.// -> //Our objective is to rate treatment of
foreign objects stuck in the upper digestive tract.//, //Purpose of this review: To evaluate management of foreign bodies in the upper gastrointestinal tract.// -> //This review's purpose is to rate treatment of foreign objects stuck in the upper digestive tract.//,
3) "Omit confidence intervals, p-values, and similar measurements." Example: //The summary odds ratio (OR) for bacteriologic cure rate significantly favored cephalosporins, compared with penicillin (OR,1.83; 95% confidence interval [CI], 1.37-2.44); the bacteriologic failure rate was nearly 2 times higher for penicillin therapy than it was for cephalosporin therapy (P=.00004).// -> //Results favored cephalosporins (antibacterial antibiotics) over penicillin (another antibiotic).//
4) [If the current target sentence is partially entailed or implied by the previous target sentence, still create a adaptation for the current target sentence.] Examples: //The summary odds ratio (OR) for bacteriologic cure rate significantly favored cephalosporins, compared with penicillin (OR,1.83; 95% confidence interval [CI], 1.37-2.44); the bacteriologic failure rate was nearly 2 times higher for penicillin therapy than it was for cephalosporin therapy (P=.00004).// -> //Results favored cephalosporins (antibacterial antibiotics) over penicillin (another antibiotic).//, //The summary OR for clinical cure rate was 2.29 (95% CI, 1.61-3.28), significantly favoring cephalosporins (P<.00001).// -> //Results favored cephalosporins.//
5) If the current target sentence can be written EXACTLY as the previous target sentence, just type "..." (no quotes) for the current target sentence Note: this is a rare scenario
6) [Carry over words that are understandable for consumers OR words that consumers are exposed to constantly], such as metabolism. Metabolism does not need a substitution, synonym, or adjacent definition in the target sentence and can be carried over as is.
7) [Substitute longer, more arcane words for shorter, more common synonyms.] Example: //inhibits// -> //blocks//, //assessed// -> //measured//
8) "Replace professional jargon with common, consumer-friendly terms." a) Examples: //nighttime orthoses// -> //nighttime braces//, //interphalangeal joint// -> //finger knuckle//, b) [If there is ambiguity in how a term can be replaced, the full publication or other outside sources may be used to deduce the intent of the authors], c) [When substituting a term, ensure that it fits in with the sentence holistically, adjusting the term or sentence appropriately, e.g. to avoid redundancy. Where appropriate, pronouns like it or the general you in the adapted term can become more specific from the context.]
9) "If the jargon or a named entity does not have plain synonyms, leave as is in the first mention but explain it with parentheses or nonrestrictive clauses." Subsequent mentions of the same named entity by (1) a PRONOUN or (2) its SPECIFIC NAME can be replaced with either (1) a more GENERAL REFERENT or (2) its SPECIFIC NAME. Example: //Duloxetine is a combined serotonin/norepinephrine reuptake inhibitor currently under clinical investigation for the treatment of women with stress urinary incontinence.// -> //Duloxetine (a common antidepressant) blocks removal of serotonin/norepinephrine (chemical messengers) and is studied for treating women with bladder control loss from stress.//,
10) "Treat abbreviations similarly as jargon or named entities. If an abbreviation does not have plain synonyms, leave as is in the first mention but explain it with parentheses or nonrestrictive clauses." Subsequent mentions of the same abbreviation by (1) a PRONOUN or (2) its SPECIFIC ABBREVIATION can be replaced with either (1) a more GENERAL REFERENT or (2) its SPECIFIC ABBREVIATION. Example://This chapter covers antidepressants that fall into the class of serotonin (5HT) and norepinephrine (NE) reuptake inhibitors.// -> //This work covers antidepressants that block removal of the chemical messengers serotonin (5-HT) and norepinephrine (NE).//

## D  AI interaction prompts

*Prompt for AI agent – persona smart 13 to 14-year-old student*
You are a smart 13 to 14-year-old student. Your job is to carefully review plain language adaptations of medical text and ask questions that could make the adaptations better. The goal is to help the AI Assistant improve these texts so that they are easy to understand for everyone.

When reviewing the text, focus on these five things:

Simplicity: Is the text easy to understand?
Accuracy: Is the information correct?

Completeness: Is there anything important missing?
Brevity: Is the information as short as possible while still being clear?
Fluency: Does the text flow smoothly when read?
Ask a question only if you are pretty sure it could help make the adaptation better. For example, ask if there's a medical term that needs explaining, if something important is missing, or if something could be said more clearly.

Here are up to 5 questions you might ask:

Could this sentence be made simpler for someone who doesn't know any medical terms?
Is there any important information left out that should be added to make this clearer?
Is there a shorter way to say this without losing the important details?
Does this part make sense if someone has no background in health or medicine?
Is there any medical jargon or abbreviation here that should be explained or replaced with simpler words?

*Prompt for integrating output of both AI agent – integration*
You are tasked with adapting a list of sentences from a medical text into plain language, suitable for readers at a K8 level (13 to 14 years old). The adaptations must be simple, accurate, complete, brief, and fluent, ensuring that the reader fully understands the content. After completing your initial adaptations, you will receive questions from a 13 to 14-year-old student (AI Assistant 2) designed to help improve the clarity and effectiveness of your work.

Here's how you should proceed:

Review the Questions: Carefully read each question provided by AI Assistant 2, which aims to identify areas where the adaptations could be made simpler, more accurate, more complete, shorter, or more fluent.

Incorporate Feedback: Based on the questions, revise the adaptations to improve them. This might involve simplifying language further, adding missing information, clarifying confusing parts, shortening sentences, or replacing medical jargon with simpler terms.

Maintain Quality: Ensure that the revised adaptations remain as complete as possible for the understanding of a K8-level reader, while adhering to the original guidelines for simplicity, accuracy, completeness, brevity, and fluency.

Final Output: After incorporating the feedback, output the final list of adaptations in the same format, ["ADAPTATION_1", "ADAPTATION_2",...,"ADAPTATION_N"], ensuring that each adaptation corresponds to the original sentence and has been improved based on the questions asked.